\pdfoutput=1

\documentclass[11pt]{article}

\usepackage{EACL2023}

\usepackage{times}
\usepackage{latexsym}

\usepackage[T1]{fontenc}

\usepackage[utf8]{inputenc}

\usepackage{microtype}

\usepackage{inconsolata}

\usepackage{graphicx}
\usepackage{todonotes}

\usepackage{svg}

\title{Modelling Temporal Document Sequences for Clinical ICD Coding}

\author{Clarence Boon Liang Ng\textsuperscript{1} \\\And
  Diogo Santos\textsuperscript{2} \\\And
  Marek Rei\textsuperscript{1,2} \AND
  \normalfont{\textsuperscript{1} Imperial College London, United Kingdom} \\
  \textsuperscript{2} Transformative AI, United Kingdom \\
  \texttt{\{clarence.ng21,marek.rei\}@imperial.ac.uk} \\
  \texttt{\{santos\}@transformative.ai}
  }

\begin{document}
\maketitle
\begin{abstract}
Past studies on the ICD coding problem focus on predicting clinical codes primarily based on the discharge summary. This covers only a small fraction of the notes generated during each hospital stay and leaves potential for improving performance by analysing all the available clinical notes. We propose a hierarchical transformer architecture that uses text across the entire sequence of clinical notes in each hospital stay for ICD coding, and incorporates embeddings for text metadata such as their position, time, and type of note. While using all clinical notes increases the quantity of data substantially, superconvergence can be used to reduce training costs. We evaluate the model on the MIMIC-III dataset. Our model exceeds the prior state-of-the-art when using only discharge summaries as input, and achieves further performance improvements when all clinical notes are used as input. 
\end{abstract}

\section{Introduction}

ICD (International Classification of Diseases \cite{icd}) coding refers to the task where medical professionals classify clinical diagnoses and medical procedures associated with each patient using standardised taxonomies, which in turn supports billing, service planning and research. The process is manual and laborious in nature \cite{icderrors}, however there is potential to automate it by identifying relevant information from clinical notes, which are already captured in EHR systems. With this in mind, researchers have started to explore whether machine learning models can succeed at this task \cite{caml}. 

The current research on the ICD coding task focuses on the extraction of codes from the discharge summary. 
This document is commonly written at the end of a hospital stay and provides a textual description of the important diagnoses and procedures for a given patient, making it particularly helpful for the task.
However, many other clinical notes are also created during the hospital stay, which can provide important details or useful additional context that may be missing from the discharge summary itself.
Analysing the full sequence of notes would allow models to make more accurate decisions and make the problem more similar to a real-life setting, where clinicians have to consider all information about a patient for ICD coding, rather than information only in a single document. 

In this work we study how the inclusion of clinical notes across the entire hospital stay can affect performance on the ICD coding task.  We propose the \textbf{H}ierarchical \textbf{T}ransformers for \textbf{D}ocument \textbf{S}equences (\textbf{HTDS}) model, which is an adaptation of the hierarchical transformer model \cite{hibert} for temporal modelling of document sequences. The model takes text and metadata (such as the time and type of note) from a sequence of multiple documents as input and achieves improved performance when additional clinical notes are used for modelling. We compare different prioritisation criteria for selecting which notes to use as input and how to best represent the sequence information. Methods related to superconvergence are applied to speed up the model training process in order to handle the increased size of the data that needs to be processed. 

Our experiments show that the inclusion of additional clinical notes indeed improves model accuracy and leads to better predictions.
We evaluate our models against the MIMIC-III-50 \cite{mimic} test set. When considering only the discharge summaries of each hospital stay as input, our model exceeds the current state-of-the-art performance in terms of Micro-F1. When considering all clinical notes as input, further performance improvements across all metrics of interest are observed, exceeding the state-of-the-art performance in Micro-F1, Micro-AUC, Macro-AUC, and Precision@5 scores. 

\section{Related Work}

Publicly available electronic health record (EHR) datasets, such as the Medical Information Mart for Intensive Care III (MIMIC-III) dataset \cite{mimic}, provide a shared context for researchers to work on ICD coding. Recent work on ICD coding concentrates on the benchmark tasks presented by \citet{caml}, which extracts ICD codes from the free-text discharge summary generated at the end of each hospital stay. \citet{caml} also publicly release their data preprocessing codes and train/dev/test data splits, and these were followed by later works for comparability of result. 

In recent years, state-of-the-art work on the ICD coding problem commonly used methods based on convolutional neural networks (CNNs) or recurrent neural networks (RNNs) for text encoding. CAML \cite{caml} uses a single convolutional layer along with “per-label attention” to extract representations for each label from the convolution output. MSAttKG \cite{msattkg} improves the performance further by using a densely connected convolutional network with variable n-gram features, and incorporating knowledge graphs to capture relationships between medical codes. EffectiveCAN \cite{effectivecan} uses a deep convolutional approach, with a “squeeze-and-excitation” module that repeatedly compresses and then decompresses the convolutional features. LAAT \cite{laat} uses a bidirectional LSTM to encode the texts, with a per-label attention step on the output to get the final classification. MSMN \cite{msmn} uses the same architecture as LAAT, with an additional step of extending code descriptions from the Unified Medical Language System (UMLS) with synonyms, and using an attention layer with a separate head for each code synonym. 

Researchers using transformer-based models for text encoding experienced difficulties in matching state-of-the-art performance. \citet{ji} apply a range of different transformer-based models but found that none of them outperformed their reimplementation of a simple CNN-based model. \citet{pascual} similarly found it difficult to achieve competitive performance and concluded that better methods of handling long input sequences are required to improve the models further. \citet{gao} also find that a simple self-attention network with far less parameters outperformed BERT-based models on many tasks. \citet{trldc} show that incorporating task-adaptive pre-training, overlapping chunks, and using a large pretrained language model make it possible to achieve performance that is close to, but still slightly below the state-of-the-art. In general, language models that were pretrained on texts in the biomedicine domain, such as ClinicalBERT \cite{clinicalbert}, BioBERT \cite{biobert}, BlueBERT \cite{bluebert}, and PubMedBERT \cite{pubmedbert} tend to achieve higher performance \cite{trldc, ji} as compared to language models such as BERT \cite{bert} and RoBERTa \cite{roberta} which are trained on general domain corpora, as the models have been adapted to the specialised language used in clinical notes. Among the range of pretrained language models available for the biomedicine domain, better performance was achieved when a specialised token vocabulary is used \cite{pubmedbert,domainroberta} and when the pre-training corpora is closer in nature to those used for the downstream task \cite{dontstoppretraining}. Very recently, \citet{plmicd} identified the restricted capacity of the [CLS] token as a potential limiting factor, and showed how using all tokens in the label attention step leads to state-of-the-art performance on the MIMIC-III-Full problem. However, they do not report results on the MIMIC-III-50 problem.  

While transformer-based language models have been very successful on short sequences of text (BERT \cite{bert} and RoBERTa \cite{roberta} use a maximum sequence length of 512 tokens), challenges arise when attempting to apply it to longer text sequences due to the quadratic computational complexity of the self-attention mechanism. Experiments conducted by \citet{gao} show that transformer models require 3x more processing time compared to CNNs, making it more tedious to explore different hyperparameters and modelling strategies. Various modifications have been proposed to the transformer architecture to reduce computation costs, in models such as TransformerXL \cite{transformerxl}, LongFormer \cite{longformer}, and BigBird \cite{bigbird}, however domain-pretrained models for these architectures are relatively scarce. Most transformer-based models for the ICD coding problem adapt the hierarchical transformer \cite{hibert}, which splits the text into chunks that are encoded separately with the pre-trained language model, and then feeds the output of the [CLS] token into a second transformer to allow interaction of information across chunks.

To the best of our knowledge, there has been no prior work that attempts to extend the ICD coding task with other clinical documents. 

\section{Approach}

\begin{figure*}[h]
    \centering
    \includegraphics[scale=0.7]{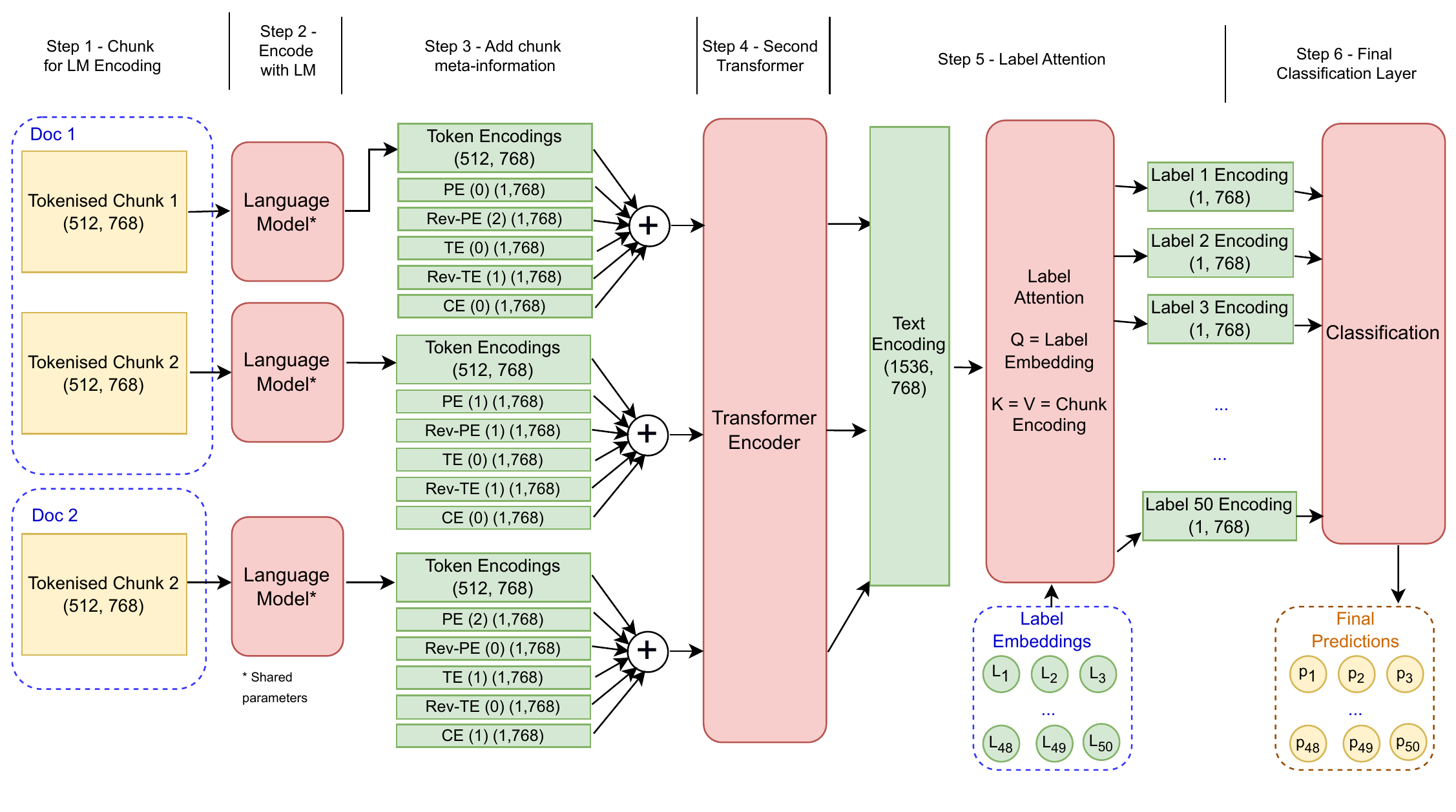}
    \caption{HTDS Model Architecture. The document sequence is first split into chunks (Step 1) and encoded with a pre-trained language model (Step 2). Meta-information of each chunk is then added to the token encodings (Step 3) before a second transformer is applied to allow attention of information across chunks (Step 4). Finally a label attention layer is applied (Step 5) and the outputs are used for classification (Step 6).}
    \label{fig:architecture}
\end{figure*}

Our \textbf{H}ierarchical \textbf{T}ransformers for \textbf{D}ocument \textbf{S}equences (\textbf{HTDS}) model is based on the hierarchical transformer architecture \cite{hibert}, with additional adaptations specifically to handle document sequences. Figure \ref{fig:architecture}
provides an illustrated diagram of the full HTDS model architecture. We process documents using the following steps: 

\textbf{Step 1 - Preprocess and Chunk}: The text in each document is sequentially tokenized and split into chunks, each containing up to $T_c$ tokens. Every new document or clinical note always starts a new chunk. 

From these tokenized chunks we select up to $N_c$ chunks for processing. If more than $N_c$ chunks are available, various prioritisation strategies can be considered to select which chunks to use as model input. In our main model we use a strategy that prioritized diversity in the categories of notes used. To do this, we select the last note by timestamp of each category, and then the second last note of each category, and so on until $N_c$ chunks of text are selected.  

\textbf{Step 2 - Encode with Language Model}: The chunks are encoded using the pre-trained language model, producing an output of dimension $N_c$ x $T_c$ x $H_e$, where $H_e$ is the dimension of the hidden state in the pre-trained LM. 

\textbf{Step 3 - Add Chunk Meta-Information}: Meta-information of each chunk is added. These are learnable embeddings, retrieved via index lookup, with size $H_e$. Positional Embeddings (PE) capture the positional index of the chunk, and are numbered from 0 for the first chunk until N-1 for the last chunk. Temporal Sequence Embeddings (TE) capture the temporal order in which the documents were captured, and are indexed in running order from 0 for chunks belonging to the first document and incremented with each subsequent document. We noted that this indexing approach would often assign varying indices to the last chunk or document, as the number of chunks and documents for each case would vary. This might limit the ability of the model to identify the last chunk or document of the text. Hence, we also include Reversed Positional Embeddings (Rev-PE) and Reversed Temporal Sequence Embeddings (Rev-TE), which start from 0 for the last chunk (or document) and are then incremented with each preceding chunk (or document). Category Embeddings (CE) capture the category of the note, with a unique index for each CATEGORY code. All learnable embeddings use values initialised from a $N(0, 0.1)$ distribution. We hypothesise that these embeddings can help the model to factor in chunk meta-information which may be relevant for classification. 

\textbf{Step 4 - Second Transformer}: The embeddings are added together (token embeddings + meta-information embeddings), then concatenated across all the chunks and given as input to a second transformer with $N_e$ encoder layers.
This allows for information from each chunk to interact with all the other chunks and the use of only a small number of layers in this second transformer will keep the computational requirements feasible. The output is an updated embedding of each token, with dimensions ($N_c$ x $T_c$) x $H_e$.

\textbf{Step 5 - Label Attention}: A label attention layer is applied. We train learnable embeddings $\alpha_l$ for each label ($\alpha = [\alpha_1 ... \alpha_{N_l}]$ has dimensions $N_l$ x $H_e$, where $N_l$ is the number of labels) which are applied against the chunk embeddings ($H = [h_1 ... h_{N_c}]$) in an attention step as follows: 

\begin{center}
    $$A = softmax(H\alpha^T)$$
    $$V = H^T A$$ 
    $$Dim(A) = (N_c \times T_c) \times N_l$$ 
    $$Dim(V) = H_e \times N_l$$
\end{center}

The i-th column in V would be an independent representation, of dimension $H_e$, for the i-th label for classification.

\textbf{Step 6 - Generate Final Classification}: A classification layer is applied. We take $\sigma(W_l v_l)$ to get the probability of label $l$, where $W_l$ is a learnable weight matrix of dimension $H_e$ for label $l$, $v_l$ is the $l$-th item of matrix V, and $\sigma$ is the sigmoid activation function. To obtain the final classification we apply a threshold $t$ for positive classification that is optimised for micro-F1 on the validation set. 

\section{Experiment Setup}

\begin{table}[t]
\centering
\begin{tabular}{lcc}
\hline \textbf{} & \textbf{Mean} & \textbf{SD}  \\ \hline
\emph{Discharge Summaries} \\
Total Documents & 1.1 & 0.4 \\
Total Words & 1896 & 929 \\
Total Tokens & 3594 & 1760 \\ 
{} \\
\emph{All Notes}\\
Total Documents & 33 & 59 \\
Total Words & 10442 & 21334 \\
Total Tokens & 21916 & 46461 \\
\hline
\end{tabular}
\caption{\label{table:summarystats} Summary statistics: Amount of text contained in clinical documents per hospital stay, measured in terms of total number of documents, words, tokens (using the RoBERTa-PM-M3-Voc tokenizer).}
\end{table}

\textbf{Dataset}: For our experiments, we use the MIMIC-III (Medical Information Mart for Intensive Care) dataset \cite{mimic}, which contains multi-modal information on patients admitted to critical care between 2001 and 2012 at the Beth Israel Deaconess Medical Center in Boston, Massachusetts. To limit computational costs, we focus on the MIMIC-III-50 problem, which limits the problem to the top 50 ICD codes by frequency. 

To construct the task dataset, we follow \citet{caml} preprocessing steps, with a few exceptions: (1) we keep the text and metadata (specifically the datetime and the category of note) of all notes rather than just the discharge summaries, (2) we do not remove punctuation as we found that performance drops when punctuation is excluded. Each record represents one hospital stay (uniquely identified by the HADM\_ID value) and  contains the texts and ICD codes linked to that hospital stay. There are 8066, 1573 and 1729 records in the train, dev and test sets respectively, giving us a total of 11368 records.

During the data cleaning process, we noticed that the train set contains clinical notes tagged under the category "Nursing/Other", but no clinical notes were tagged in this category in the dev and test sets. For our experiments we grouped "Nursing/Other" and "Nursing" into a single category. 

Table \ref{table:summarystats} shows summary statistics of the dataset. In general, discharge summaries are far longer than other documents, with an average of 1724 words per document as compared to the overall average of 316 words per document. However, the text in discharge summaries only accounts for less than 20\% of the words generated in each hospital stay, suggesting the possibility that the other notes might carry additional information that can improve ICD coding accuracy. We also provide the number of tokens produced when the text is tokenized with the RoBERTa-PM-M3-Voc \cite{domainroberta} tokenizer, and we see from the numbers that most hospital stays involve text data that is beyond the 512-token maximum of a single transformer language model. 

We also note that the amount of text in each hospital stay can vary widely and has a right-skewed distribution. There is a notable proportion of longer hospital stays which generate substantially more documents and text as compared to the rest. The 90th percentile for Total Words and Total Document Count across all notes is 20556 and 72 respectively. For these hospital stays, the effects of the note prioritisation strategy on model performance would be more prominent. 

\textbf{Task Definition}: We investigate two variations of the ICD classification task on this dataset. For Task 1, the notes that are available for modelling are restricted to discharge summaries only. Some hospital stays (11\% of stays) have multiple discharge summaries, typically because of addenda, and in these cases we keep all of them. This would be equivalent to the MIMIC-III-50 task attempted by past works. For Task 2, all notes in each hospital stay are available for use in modelling. This vastly increases the number of documents (from an average of 1.1 to 33 per hospital stay) and the number of words (from an average of 1896 to 10442) to be considered. Task 2 uses the same data splits and labels as Task 1, allowing us to compare the results to assess whether information from the additional notes is able to improve performance.

For both tasks, we use the same evaluation metrics as defined by Mullenbach et al \cite{caml} and then subsequently followed by other researchers: micro-F1, macro-F1, micro-AUC, macro-AUC, and Precision at k=5.

\textbf{Implementation and Model Hyperparameters}: Pytorch was used for the implementation of our models, and NVIDIA Tesla A100 80GB GPUs were used for finetuning. Hyperparameters were tuned manually; Table \ref{table:hyperparameter_search} details the search space and final hyperparameter values used for the HTDS model. The pretrained language model was initialised to the RoBERTa-base-PM-M3-Voc \cite{domainroberta} model checkpoint, which was pretrained on texts in PubMed, PubMed Central, and MIMIC-III physician notes. The second transformer uses 1 encoder layer with 8 attention heads. 

\begin{table}[t]
\begin{tabular}
{p{3.3cm}p{3.3cm}}
\hline  Hyperparameter & Values\\
\hline
\emph{Optimization} \\
Peak Learning Rate & 1e-6 to 1e-4 \textbf{(5e-5)}\\
Number of Epochs & 10-50 \textbf{(20)} \\
Early Stopping Patience Threshold & None, \textbf{3}, 5, 10 \\
Effective Batch Size & 1-64 \textbf{(16)} \\ 
\\
\emph{Language Model} \\
Pre-trained LM & PubMedBERT, \textbf{RoBERTa-base-PM-M3-Voc}, RoBERTa-large-PM-M3-Voc\\
Tokens per chunk, $T_c$ & \textbf{512}\\
Max Chunks, $N_c$ & 1-48 \textbf{(32)}\\
\\
\emph{Second Transformer} \\
Encoder Layers & 0, \textbf{1}, 2 \\ 
Attention Heads & \textbf{8}, 12 \\
\hline
\end{tabular}
\caption{\label{table:hyperparameter_search} Hyperparameter search space. Bolded text indicates hyperparameters used in the HTDS model.}
\end{table}

Texts are tokenized into chunks of $T_c$=512 tokens and a maximum of $N_c$=32 chunks were used as model input. With these values for $T_c$ and $N_c$, the note selection strategy to maximise diversity of document categories (detailed earlier in Section 3) was applied for 45\% of samples which have more than 32 chunks of text. The model has 136M parameters in total.

\begin{table*}[h]
\begin{center}
\begin{tabular}
{lccccc}
\hline  {} & Micro $F_1$ & Macro $F_1$ & Micro AUC & Macro AUC & P@5 \\
\hline 
\multicolumn{2}{l}{\emph{CNN-based Models}} \\ 
CAML \cite{caml} & 63.3 & 57.6 & 91.6 & 88.4 & 61.8 \\
MSAttKG \cite{msattkg} & 68.4 & 63.8 & 93.6 & 91.4 & 64.4 \\
EffectiveCAN \cite{effectivecan} & 71.7 & 66.8 & 94.5 & 92.0 & 66.4 \\
\hline 
\multicolumn{2}{l}{\emph{RNN-based Models}} \\ 
LAAT \cite{caml} & 71.5 & 66.6 & 94.6 & 92.5 & 67.5 \\
MSMN \cite{msmn} & 72.5 & \textbf{68.3} & 94.7 & 92.8 & 68.0 \\
\hline 
\multicolumn{2}{l}{\emph{Transformer-based Models}} \\ 
Hier-PubMedBERT \cite{ji} & 68.1 & 63.3 & 90.8 & 88.6 & 64.4 \\
TrLDC (Base) \cite{trldc} & 70.1 & 63.8 & 93.7 & 91.4 & 65.9 \\
TrLDC (Large) \cite{trldc} & 71.1 & 65.5 & 94.1 & 91.9 & 66.4 \\
\hline 
\multicolumn{2}{l}{\emph{Our Models}} \\ 
HTDS (Discharge Summaries) & $72.6_{0.3}$ & $66.6_{1.2}$ & $94.5_{0.1}$ & $92.6_{0.3}$ & $67.4_{0.3}$ \\
HTDS (All Notes) & \textbf{73.3}$_{0.3}$ & $67.9_{0.4}$ & \textbf{95.0}$_{0.2}$ & \textbf{93.2}$_{0.2}$ & \textbf{68.2}$_{0.2}$ \\
\hline
\end{tabular}
\caption{\label{table:main_results} Performance of models on the MIMIC-III-50 test set. Models are sorted by Micro-F1 within each category. Metrics are averaged across 5 replications. Subscripts indicate the standard deviation across runs. Bolded values indicate the best score achieved for each metric.}
\end{center}
\end{table*}

These hyperparameters were selected to maximise Micro-F1 on the dev set, with a few exceptions to manage training and computation costs: (1) while using the larger RoBERTa-large-PM-M3-Voc model was found to achieve better performance, we kept to the smaller RoBERTa-base-PM-M3-Voc model; (2) while increasing the maximum number of chunks $N_c$ in general leads to better performance, we limit our model to a maximum of 32 chunks. 

Training models that take text across all clinical documents as inputs, compared to using only the discharge summary, requires substantially more computational resources. With A100 GPUs, 15.5 samples are processed per second when training on discharge summaries only\footnote{TrLDC \cite{trldc}, which we consider to be a comparable model in terms of architecture, processed 7.4 samples per second when training on discharge summaries on NVIDIA V100 GPUs.}, and 4.9 samples are processed per second when training with all clinical documents.
To speed up the model optimisation process, we apply the 3-phase \emph{1cycle}
learning rate scheduler for superconvergence as described in \cite{superconvergence}. The learning rate (LR) progresses via cosine annealing from 1/25 of the peak LR to the peak LR (5e-5) in the first phase (30\% of iterations) and then goes back to 1/25 of the peak LR in the second phase (30\% of iterations). Finally in the third phase (40\% of iterations), LR is annealed to 1/1000 of the peak LR. The AdamW optimizer is used, with an effective batch size of 16 achieved through gradient accumulation. The model is trained for up to 20 epochs with an early stopping patience threshold of 5. With this setup, training is stopped at around the 14th epoch on average. We note that this is at least 50\% less (in terms of number of epochs) compared to past works on the MIMIC-III-50 problem where transformer-based models would be trained for 30 epochs or more \cite{trldc, ji, pascual}.

\section{Results}

\subsection{Main Results} Table \ref{table:main_results} shows the results when our models are evaluated against the MIMIC-III-50 test set, as well as comparisons against published works. We report the averaged metrics across 5 training replications. 

As we can see from the table, prior works with transformer-based models faced challenges in achieving competitive performance on this problem. \citet{trldc} managed substantial improvements with the TrLDC model over the work of \citet{ji}, however even with a large-sized model their performance still fell slightly behind the best-performing CNN-based and RNN-based models. When using only discharge summaries, HTDS achieves state-of-the-art performance in terms of Micro-F1, the primary metric used for comparison. It also exceeds past CNN-based and Transformer-based models on all metrics of interest. 

When including all clinical documents, as compared to including only discharge summaries, the performance of HTDS improves on all metrics of interest (all differences are statistically significant at p<0.05), including an additional 0.7\% increase in Micro-F1. Comparing against all other models, we see that the model achieves state-of-the-art performance in terms of all metrics except for Macro-F1. We hypothesize that the modelling of code synonyms in MSMN \cite{msmn} helped to increase its performance on rarer ICD codes and hence achieve a higher Macro-F1 score, but also note that steps used to improve performance by incorporating synonyms based on UMLS concepts could also be adapted into our model to achieve similar improvements.

Put together, our results demonstrate the value of including clinical documents beyond the discharge summary in modelling. 

\subsection{Ablation Experiments}
To analyse the effect of various components and hyperparameter choices on model performance, we start with our main model and then ablate or vary individual components one at a time, keeping all other components constant, and evaluate their performance on the development set. We share our results in this section. 

For all ablation experiments, we report the impact on Micro-F1, the primary metric of interest, averaged across 5 replications. 

\textbf{Quantity of Text Input}:
Table \ref{table:qty_input} shows how performance varies as the quantity of text is varied. The quantity of text used as input has a substantial impact on the compute requirements of the entire model architecture. When $N_c$ is reduced 16, 7.5 samples are processed per second when training on A100 GPUs, an increase of 0.5x as compared to 4.9 samples per second for HTDS which uses $N_c$=32. However, as we can see from the results of this ablation experiment, reducing the quantity of text input leads to a substantial drop in model performance.

\begin{table}[h]
\begin{tabular}
{p{5.1cm}cc}
\hline  {} & Micro $F_1$\\
\hline
HTDS (Max 32 Chunks) & 74.0\\
Max 16 Chunks & 73.0\\
\hline
\end{tabular}
\caption{\label{table:qty_input} Performance when the quantity of text input is varied on the development set.}
\end{table}

\textbf{Metadata embeddings}: Table \ref{table:metadata embeddings} shows how the performance varies as the metadata embeddings used in the model are varied. The ablation of each of the embedding types in isolation results in small but consistent decreases in model performance. It is possible that the model compensates by learning to capture some of this information from the text itself without explicit embeddings. Indeed, past works have observed that the clinical notes in the MIMIC-III dataset have a high degree of structure and templating \cite{ehrwriting}. Nevertheless, in our experiments the overall best results were obtained by using the combination of all the proposed metadata embeddings.

\begin{table}[h]
\begin{center}
\begin{tabular}
{p{5.1cm}ccccc}
\hline  {} & Micro $F_1$\\
\hline
HTDS (All meta embeddings) & 74.0 \\
Ablate CE & 73.9 \\
Ablate PE+Rev-PE & 73.9\\
Ablate TE+Rev-TE & 73.8\\
\hline
\end{tabular}
\caption{\label{table:metadata embeddings} Performance when metadata embeddings are varied on the development set. }
\end{center}
\end{table}

\textbf{Chunk Representations}: In a traditional hierarchical transformer, only the encoding of the [CLS] token is kept and used as an aggregate representation of the chunk. However, recent works have suggested that the [CLS] token might have insufficient capacity to capture information for the large number of labels in the ICD coding problem \cite{plmicd}. In Table \ref{table:chunk_representation}, we show the results when only the [CLS] token is used as an aggregate representation of each chunk, and see that there is a sizeable decrease in performance. 

\begin{table}[h]
\begin{center}
\begin{tabular}
{p{5.1cm}ccccc}
\hline  {} & Micro $F_1$\\
\hline
HTDS (All token representations) & 74.0 \\
CLS token representation only & 71.7 \\
\hline
\end{tabular}
\caption{\label{table:chunk_representation} Performance when the embeddings used for chunk representation are varied on the development set. }
\end{center}
\end{table}

\textbf{Second Transformer}: The second transformer in Step 4 allows tokens from each chunk to attend to tokens from other chunks. While earlier studies \cite{trldc,ji} include this second transformer, it also adds to the computational costs of the model due to the quadratic complexity of the attention step and \cite{plmicd} show that the second transformer can be dropped if the encodings of all tokens (rather than just the [CLS] token) are kept for the label attention step. 

Our ablation experiments in Table \ref{table:second_transformer} provide some additional insight on this. When considering only the discharge summary, the second transformer can be dropped without substantial impact on performance. However, when modelling the sequence of all clinical documents, ablating the second transformer leads to a noticeable decrease in performance, suggesting that the information in other documents can help further refine token representations before classification.  

\begin{table}[h]
\begin{center}
\begin{tabular}
{p{5.1cm}ccccc}
\hline  {} & Micro $F_1$ \\
\hline
HTDS (Discharge Summaries) & 73.2 \\ 
Ablate 2nd Transformer & 73.3 \\ 
\\
HTDS (All Notes) & 74.0 \\
Ablate 2nd Transformer & 73.6 \\ 
\hline
\end{tabular}
\caption{\label{table:second_transformer} Performance when second transformer is ablated on the development set.}
\end{center}
\end{table}

\textbf{Note Selection:}
In around 45\% of admissions, tokenizing the text in all available clinical notes will produce more than 32 chunks of 512 tokens. In those cases, we would need to select which chunks are used as inputs to our model. Table \ref{table:note selection} shows our results. We considered the following strategies to prioritise which chunks to use: 
\begin{itemize}
    \item By timestamp: We select the first or last 32 chunks by the timestamp of the clinical notes. Taking the last chunks achieved far superior performance. 
    \item By category: We select first the discharge summary\footnote{Our exploratory tests find that the discharge summaries contain the most relevant information. We note also that prior work achieved good performance with just the discharge summaries, without the need for other notes.}, then notes of a certain category (Radiology/Nursing/Physician), and then other notes until 32 chunks of text are selected. Our results indicate that the differences in performance are mostly marginal, suggesting that there could be multiple possible strategies that achieve close to optimal performance.
    \item Prioritise diversity:  We select first the last note by timestamp of each category, and then the second last note of each category, and so on until 32 chunks of text are selected. This maximises the diversity (in terms of categories of notes) used as inputs to the model. This approach was found to have the highest score on the development set, and hence used for HTDS. 
\end{itemize}

\begin{table}[h]
\begin{center}
\begin{tabular}
{p{5.1cm}cc}
\hline  {} & Micro $F_1$ \\
\hline
HTDS (Prioritise diversity) & 74.0 \\
\\
Prioritise First & 68.4\\
Prioritise Last & 73.8 \\
\\
Prioritise Radiology & 73.8 \\
Prioritise Nursing & 73.7 \\
Prioritise Physician & 73.8 \\
\hline
\end{tabular}
\caption{\label{table:note selection} Performance when note selection is varied on the development set. }
\end{center}
\end{table}

In general, we also note that the effects of note selection strategies would be more pronounced when the maximum number of chunks $N_c$ for model input is smaller, as it would result in a greater proportion of text being excluded. 

\section{Conclusion}

As we work towards automated ICD coding, it would be helpful to build models that can consider information that is captured across the patient's EHR record, rather than just the discharge summary (which may not always be exhaustive). Such an approach would also be more similar to a real-life setting, where clinicians consider all available information for ICD coding, rather than information in a single document.

In this paper, we demonstrated the HTDS model, which is an adaptation of the hierarchical transformer model that considers the text and metadata from the entire clinical document sequence for ICD coding. While transformer-based models have faced difficulties achieving competitive performance on the ICD coding problem in the past, with HTDS we show that these challenges can be overcome. When evaluated on the MIMIC-III-50 test set using only discharge summaries, HTDS exceeded the prior state-of-the-art performance in terms of Micro-F1 score. When all clinical documents were considered, the performance of HTDS improved further on all performance metrics of interest, and exceeded prior state-of-the-art performance in all metrics except for Macro-F1. The results demonstrate the value of including clinical documents beyond the discharge summary in the ICD coding problem.

Possibilities for improving performance even further are plenty. These include: using a large-sized language model or using overlapping text chunks to reduce fragmentation in the initial encoding step \cite{trldc}, considering other transformer architectures for long texts \cite{longformer, transformerxl, bigbird}, smarter strategies for chunking the input text to reduce fragmentation, further improving the strategy for selecting which text to use as model input (possibly going down to text-level rather than document-level approaches), and incorporating methods to better model rare ICD codes \cite{laat, msmn}. Approaches for improving the computational efficiency and training time of the model can be considered to help to reduce GPU resource requirements, and enable the testing of more models and hyperparameter settings. Going even further from here, we could consider multi-modal models that use information across the entire EHR database for ICD coding. 

We hope that our findings will encourage future studies that tap on the full breadth and depth of information available in modern EHR databases today in order to further push the limits of performance on the ICD coding problem in future. 

\section*{Limitations}
Although applying HTDS on the full clinical document sequence in each hospital stay helped to push performance on the ICD coding problem further as compared to the prior state-of-the-art, we note a few limitations to our work. 

Firstly, the computational requirements to train HTDS is not trivial. When using NVIDIA A100 GPUs, one training run took 8 GPU-hours on average (for 5 replications this would require 40 GPU-hours). The increased computation cost for HTDS, as compared to other models on the ICD coding problem, could be attributed to the higher number of model parameters in transformers as compared to CNN/RNNs and the increase in input data size as a result of using all clinical documents as input. It is hoped that this issue of high compute costs can be mitigated in future by either further refinements in modelling to improve efficiency or improvements in the compute capabilities of hardware used for model training.  

Secondly, we note that our work focuses only on the MIMIC-III-50 problem, where only the top 50 ICD codes by frequency are considered. This would be insufficient in a real-life setting, which would require clinicians to consider all ICD codes. Extending our work on the MIMIC-III-Full problem, which uses a dataset that is 4x in size, was not attempted due to limitations on compute resources. However, we speculate that the benefits of using all clinical documents to perform ICD coding would apply similarly to the MIMIC-III-Full problem. 

Finally, while we have taken the actual ICD codes assigned by clinicians as the "ground truth" for the purpose of model evaluation, there have been errors made during the process. We would not expect clinicians to thoroughly read the entire clinical document sequence (consisting an average of over 10,000 words) for every patient when performing ICD coding, and hence there is a possibility that some codes could have been missed. A more thorough approach for model evaluation could involve extracting a sample of records where different codes were assigned by the clinicians and our models for further evaluation by experts, in order to determine the extent to which this might have affected our evaluation metrics. 

\section*{Ethics Statement}
No conflicts of interest are declared by the authors. Clinical data in the MIMIC-III database is de-identified through removal of identifying data elements and date-shifting in accordance with Health Insurance Portability and Accountability Act (HIPAA) standards, and protected health information was further removed from clinical notes using dictionary look-ups and pattern-matching \cite{mimic}. The use of the data is in accordance with the MIMIC-III data use agreement.

\bibliography{anthology,custom}
\bibliographystyle{acl_natbib}

\appendix



\end{document}